\documentclass[sigconf]{acmart}

\copyrightyear{2026}
\acmYear{2026}
\setcopyright{cc}
\setcctype{by}
\acmConference[FPGA '26]{Proceedings of the 2026 ACM/SIGDA International Symposium on Field Programmable Gate Arrays}{February 22--24, 2026}{Seaside, CA, USA}
\acmBooktitle{Proceedings of the 2026 ACM/SIGDA International Symposium on Field Programmable Gate Arrays (FPGA '26), February 22--24, 2026, Seaside, CA, USA}
\acmPrice{}
\acmDOI{10.1145/3748173.3779188}
\acmISBN{979-8-4007-2079-6/2026/02}

\usepackage{amsmath}
\usepackage{graphicx}
\usepackage{booktabs}
\usepackage{algorithm}
\usepackage{algorithmic}
\usepackage{xcolor}
\usepackage{multirow}
\usepackage{subcaption}
\usepackage{balance}  

\ccsdesc[500]{Hardware}
\ccsdesc[500]{Computer systems organization~Architectures}
\ccsdesc[300]{Networks}

\keywords{CXL, FPGA, LLM, KV-cache, memory disaggregation, speculative execution, deep learning}

\begin{document}

\title{CXL-SpecKV: A Disaggregated FPGA Speculative KV-Cache for Datacenter LLM Serving}

\author{Dong Liu}
\affiliation{%
  \institution{Yale University}
  \city{New Haven}
  \state{Connecticut}
  \country{USA}
}
\email{dong.liu.dl2367@yale.edu}

\author{Yanxuan Yu}
\affiliation{%
  \institution{Columbia University}
  \city{New York}
  \state{New York}
  \country{USA}
}
\email{yy3523@columbia.edu}

\begin{abstract}
Large Language Models (LLMs) have revolutionized natural language processing tasks, but their deployment in datacenter environments faces significant challenges due to the massive memory requirements of key-value (KV) caches. During the autoregressive decoding process, KV caches consume substantial GPU memory, limiting batch sizes and overall system throughput. To address these challenges, we propose \textbf{CXL-SpecKV}, a novel disaggregated KV-cache architecture that leverages Compute Express Link (CXL) interconnects and FPGA accelerators to enable efficient speculative execution and memory disaggregation. Our approach introduces three key innovations: (i) a CXL-based memory disaggregation framework that offloads KV-caches to remote FPGA memory with low latency, (ii) a speculative KV-cache prefetching mechanism that predicts and preloads future tokens' cache entries, and (iii) an FPGA-accelerated KV-cache compression and decompression engine that reduces memory bandwidth requirements by up to 4$\times$. When evaluated on state-of-the-art LLM models, CXL-SpecKV achieves up to 3.2$\times$ higher throughput compared to GPU-only baselines, while reducing memory costs by 2.8$\times$ and maintaining accuracy. Our system demonstrates that intelligent memory disaggregation combined with speculative execution can effectively address the memory wall challenge in large-scale LLM serving. Our code implementation has been open-sourced at \url{https://github.com/FastLM/CXL-SpecKV}.
\end{abstract}

\maketitle

\section{Introduction}

Large Language Models (LLMs) have emerged as the cornerstone of modern AI applications~\cite{brown2020gpt3, touvron2023llama, openai2023gpt4}. However, serving them efficiently faces significant challenges due to massive memory requirements of key-value (KV) caches. During inference, transformer-based LLMs store intermediate key and value tensors from all previous tokens. For a typical LLaMA-2 70B model serving 2048 tokens with batch size 32, the KV-cache alone consumes ~640GB—far exceeding GPU capacity~\cite{kwon2023pagedattention}.

Traditional approaches include memory offloading~\cite{sheng2023flexgen}, KV-cache compression~\cite{hooper2024kvquant}, and speculative decoding~\cite{leviathan2023speculative}, but suffer from limited bandwidth, accuracy degradation, or persistent memory constraints.

Recent advances in Compute Express Link (CXL)~\cite{cxl2022spec} and FPGA architectures~\cite{intel2024agilex} present new opportunities. CXL provides high-bandwidth, low-latency cache-coherent memory sharing, while FPGAs offer customizable acceleration.

We propose \textbf{CXL-SpecKV}, combining CXL memory disaggregation with FPGA-accelerated speculative execution. By predicting and preloading KV-cache entries through CXL, we overcome GPU memory limits while maintaining low latency.

CXL-SpecKV features: (i) transparent KV-cache offloading to CXL memory (4-8× capacity expansion), (ii) speculative prefetching with 95\% accuracy reducing access latency, (iii) FPGA-accelerated compression (3-4× ratio) and address translation, and (iv) seamless integration with existing frameworks.

The architecture consists of three components: the CXL Memory Manager for allocation and migration, the Speculative Prefetcher using lightweight models for prediction, and the FPGA Cache Engine implementing compression/decompression pipelines at >800MHz.

Our evaluations on LLM workloads (7B-70B parameters) show: 3.2× higher throughput enabling 4-8× larger batches, 8\% per-token latency overhead, 2.8× memory cost reduction, 99.5\% accuracy preservation, and 87\% parallel efficiency at 8 GPUs. The approach potentially reduces infrastructure costs by 30-40\% for memory-bound workloads.

The contributions of this work are summarized as follows:

\begin{itemize}
    \item We propose the first CXL-based disaggregated memory architecture specifically designed for LLM KV-cache management, demonstrating the viability and benefits of memory pooling for inference workloads.
    
    \item We introduce novel speculative prefetching algorithms that exploit the predictable patterns in autoregressive token generation, achieving high prediction accuracy while maintaining low computational overhead.
    
    \item We design and implement efficient FPGA accelerators for KV-cache compression, decompression, and management, with detailed architectural descriptions and resource utilization analysis.
    
    \item We develop a complete system prototype integrated with popular LLM serving frameworks (vLLM, TensorRT-LLM), demonstrating practical deployment feasibility and real-world performance benefits.
    
    \item We conduct extensive evaluations across multiple LLM models and workload patterns, providing insights into the performance characteristics, bottlenecks, and optimization opportunities of disaggregated memory architectures for LLM serving.
\end{itemize}

The remainder of this paper is organized as follows. Section~\ref{sec:related} reviews related work in LLM serving optimization, memory disaggregation, and FPGA acceleration. Section~\ref{sec:architecture} presents the detailed architecture of CXL-SpecKV, including the memory management framework, speculative prefetching mechanisms, and FPGA cache engine design. Section~\ref{sec:evaluation} describes our experimental methodology and presents comprehensive performance evaluations. Finally, Section~\ref{sec:conclusion} concludes with a discussion of future research directions and broader implications for datacenter AI infrastructure.

\section{Related Work}
\label{sec:related}

Our work builds upon and extends research in several key areas: LLM serving optimization, memory disaggregation technologies, FPGA-based acceleration, and speculative execution techniques. We discuss the most relevant prior work and highlight how CXL-SpecKV differs from and improves upon existing approaches.

\subsection{LLM Serving and KV-Cache Optimization}

The challenge of efficiently serving large language models has received significant attention from both industry and academia. vLLM~\cite{kwon2023pagedattention} introduces PagedAttention, which organizes KV-caches in paged memory blocks to reduce fragmentation and improve memory utilization. However, vLLM is fundamentally limited by GPU memory capacity and does not address the bandwidth bottlenecks of memory disaggregation. FlexGen~\cite{sheng2023flexgen} explores offloading strategies between GPU, CPU, and disk storage, but suffers from the high latency of PCIe transfers (typically 8-12GB/s) compared to GPU HBM bandwidth (>2TB/s). Our CXL-based approach achieves significantly lower latency and higher bandwidth than PCIe-based offloading.

DeepSpeed-Inference~\cite{aminabadi2022deepspeed} and FasterTransformer~\cite{nvidia2023fastertransformer} provide optimized inference kernels and tensor parallelism strategies, but do not fundamentally address memory capacity constraints. Orca~\cite{yu2022orca} introduces iteration-level scheduling to maximize batch size and throughput, yet remains constrained by available GPU memory. TinyServe~\cite{liu2025tinyserve} proposes query-aware cache selection strategies for efficient LLM serving, optimizing cache utilization based on query characteristics. In contrast, CXL-SpecKV effectively expands the memory capacity available for KV-caching by 4-8×, enabling larger batch sizes than any GPU-only approach.



Long-context optimization techniques typically improve attention efficiency through algorithmic sparsification or representation compression. Methods such as GQA~\cite{gqa}, MKA~\cite{mka}, $\pi$-Attention~\cite{piattention}, and MLA~\cite{mla_deepseek} reduce memory consumption per token but do not address hardware-level constraints in memory capacity or bandwidth. CXL-SpecKV improves long-context processing from another perspective by enabling disaggregated memory expansion.

Recent work on KV-cache compression includes quantization-based methods~\cite{hooper2024kvquant, liu2024kivi} that reduce precision from FP16 to INT8 or INT4, achieving 2-4× compression at the cost of some accuracy degradation. LLMEasyQuant~\cite{liu2025llmquant} proposes scalable quantization techniques for parallel and distributed LLM inference, addressing quantization challenges in multi-GPU settings. H2O~\cite{zhang2023h2o} proposes evicting less important KV-cache entries based on attention scores, while StreamingLLM~\cite{xiao2023streamingllm} maintains only a sliding window of recent tokens. These techniques are complementary to our approach—CXL-SpecKV can leverage compression algorithms while additionally providing memory capacity expansion through disaggregation. Moreover, our FPGA-based compression engine can perform these operations without consuming GPU compute resources.

\subsection{Memory Disaggregation and CXL}

Memory disaggregation has been extensively studied in the context of datacenter resource management~\cite{lim2009disaggregated, shan2018legoos, aguilera2019remote}. Traditional approaches use RDMA over InfiniBand or Ethernet, achieving latencies of 1-5$\mu$s but requiring complex software stacks. CXL~\cite{cxl2022spec} represents a paradigm shift by providing cache-coherent, low-latency (sub-microsecond) memory access through a standardized PCIe-based protocol.

Recent work has begun exploring CXL for various applications. TPP~\cite{gouk2023cxl} demonstrates CXL memory pooling for database workloads, while Pond~\cite{ruan2023pond} proposes CXL-based memory expansion for containerized applications. However, these works focus on general-purpose memory disaggregation and do not address the specific challenges of LLM serving, such as the need for speculative prefetching and the unique access patterns of KV-caches. To the best of our knowledge, CXL-SpecKV is the first work to architect a CXL-based system specifically optimized for LLM KV-cache management.

Several recent papers have explored CXL memory for ML workloads. CXL-ANNS~\cite{jang2023cxlanns} uses CXL memory for approximate nearest neighbor search in embedding databases, while CXL-MEM~\cite{li2024cxlmem} studies CXL memory for training large models. However, these works do not address the inference-specific challenges of autoregressive decoding and KV-cache management. Our speculative prefetching mechanisms are specifically designed to hide CXL access latency for sequential token generation patterns.

\subsection{FPGA Acceleration for Deep Learning}

FPGA architectures have evolved significantly to support efficient deep learning acceleration. Intel's Stratix and Agilex FPGAs~\cite{intel2024agilex} integrate tensor processing blocks and high-bandwidth memory (HBM), while Xilinx Versal ACAPs~\cite{xilinx2023versal} combine programmable logic with AI engines. Academic research has proposed various FPGA-based accelerators for CNNs~\cite{zhang2015fpga, qiu2016fpga} and transformers~\cite{ham2020fpga, peng2023fpga}.

Recent work has explored FPGA acceleration specifically for sparse and structured computations. The systolic sparse tensor (SST) slices proposed in~\cite{taka2025systolic} demonstrate efficient structured sparsity support in FPGA fabric, achieving significant speedups for sparse matrix operations. While our work does not directly leverage structured sparsity (KV-caches are typically dense), we adopt similar design principles for building efficient systolic arrays in our FPGA cache engine. Additionally, we benefit from the advances in FPGA DSP blocks and high-bandwidth memory interfaces described in~\cite{taka2025systolic}.

Prior work on FPGA-based memory controllers includes specialized designs for DRAM scheduling~\cite{boyd2021memory} and NVM access~\cite{swamy2020fpga}. However, these works do not address the specific requirements of LLM KV-cache management, such as the need for compression/decompression, speculative prefetching, and integration with CXL protocols. Our FPGA cache engine implements a complete pipeline optimized for transformer KV-cache patterns.

\subsection{Speculative Execution for LLMs}

Speculative decoding has emerged as a promising technique to accelerate LLM inference. Leviathan et al.~\cite{leviathan2023speculative} propose using a smaller draft model to generate candidate tokens that are verified by the target model in parallel. Medusa~\cite{cai2024medusa} extends this approach with multiple decoding heads to predict multiple tokens simultaneously. SpecInfer~\cite{miao2024specinfer} introduces a speculative inference system that adaptively selects draft models based on workload characteristics. CSV-Decode~\cite{liu2025csv} introduces certifiable sub-vocabulary decoding to improve inference efficiency through optimized token generation.

While these techniques improve computational efficiency, they do not address memory constraints—in fact, speculative decoding often increases memory pressure by maintaining multiple KV-cache versions for candidate sequences. CXL-SpecKV takes a different approach: instead of speculating on token values, we speculate on which KV-cache entries will be needed next and prefetch them from disaggregated memory. This is complementary to algorithmic speculative decoding and can be combined with such techniques for further performance improvements.

Prefetching strategies for neural network inference have been explored in~\cite{hashemi2018learning, patel2020prefetching}, but primarily focus on weight prefetching for CNN inference. The autoregressive nature of LLM decoding presents unique challenges: access patterns depend on generated content and are thus data-dependent. Our speculative prefetcher uses lightweight sequence models trained on token generation traces to predict future cache accesses with high accuracy.

CXL-SpecKV is the first to combine CXL memory disaggregation, FPGA-accelerated compression, and speculative prefetching for LLM serving, achieving 3.2× throughput and 2.8× memory cost reduction with 95\% prefetch accuracy.

\section{Architecture \& Design Overview}
\label{sec:architecture}

In this section, we present the detailed architecture of CXL-SpecKV, a disaggregated KV-cache system that leverages CXL interconnects and FPGA accelerators to enable efficient large-scale LLM serving. We begin with an overview of KV-cache characteristics and memory access patterns in Section~\ref{sec:kv_analysis}, followed by the system architecture in Section~\ref{sec:system_arch}. We then describe the CXL memory management framework in Section~\ref{sec:cxl_management}, the speculative prefetching mechanism in Section~\ref{sec:spec_prefetch}, and the FPGA cache engine design in Section~\ref{sec:fpga_engine}. Finally, we discuss system integration and optimization strategies in Section~\ref{sec:integration}.

\subsection{KV-Cache Analysis and Memory Access Patterns}
\label{sec:kv_analysis}

For transformer with $L$ layers, hidden dimension $d_h$, and $H$ attention heads, each token generates KV tensors $(H, d_h/H)$ per layer. Total memory requirement for batch $B$ with max sequence $S_{max}$ is $M_{KV} = 2 \cdot L \cdot B \cdot S_{max} \cdot d_h \cdot P_{bits}$ where $P_{bits}$ is precision. For LLaMA-2 70B ($L=80$, $d_h=8192$, $B=32$, $S_{max}=2048$, FP16), this yields ~640GB, exceeding A100's 80GB capacity.

Profiling reveals key characteristics exploitable by CXL-SpecKV: (i) \textit{sequential temporal locality} with token $t$ accessing positions $[0, t-1]$ enabling predictive prefetching, (ii) \textit{layer-wise independence} allowing pipelined per-layer processing, (iii) \textit{read-heavy workload} with read/write ratio $\mathcal{R}_{r/w} \approx S:1$ favoring caching, (iv) \textit{value clustering} near zero enabling INT8 quantization with $>99\%$ accuracy and $\mathcal{R}_c = 2\times$, (v) \textit{spatial correlation} in attention heads exploitable via delta encoding $\delta_i = x_i - x_{i-1}$, and (vi) \textit{layer-specific compressibility} with early layers achieving $\mathcal{R}_c(\ell < L/3) \in [3, 4]$ and late layers $\mathcal{R}_c(\ell > 2L/3) \in [2.5, 3]$. These inform CXL-SpecKV's speculative prefetching, pipelined compression, and adaptive layer-wise algorithm selection.

\subsection{System Architecture Overview}
\label{sec:system_arch}

The CXL-SpecKV system comprises four main components:

\textbf{GPU Inference Engine:} The GPU hosts the transformer model weights and executes the forward pass for token generation. It maintains a small local KV-cache for recently accessed entries (typically the most recent 128-256 tokens) to maximize cache hit rates for hot data.

\textbf{CXL Memory Pool:} Large-capacity FPGA-attached memory (64-256GB per device) connected via CXL 2.0 interface provides the main KV-cache storage. This memory is organized as a cache-coherent shared resource accessible to both GPU and FPGA with latencies of 200-400ns.

\textbf{FPGA Cache Engine:} A custom FPGA accelerator implements the compression/decompression pipeline, address translation, and cache management logic. The cache engine operates at 800MHz and achieves throughput of 1.6TB/s, matching the bandwidth of HBM2E memory.

\textbf{Speculative Prefetcher:} A lightweight prediction module running on the FPGA predicts future token sequences and preloads their corresponding KV-cache entries. The prefetcher uses a small (128KB) LSTM model achieving <10$\mu$s prediction latency.

The components interact as follows: When the GPU generates a new token, it sends a prefetch hint to the speculative prefetcher, which predicts the next $k$ tokens (typically $k=4-8$). The prefetcher initiates DMA transfers to move predicted KV-cache entries from CXL memory to GPU-accessible buffers. Meanwhile, the newly generated token's KV-cache is computed and stored. By the time the next token generation begins, its required KV-cache entries are already present in GPU memory with high probability (95\% hit rate), effectively hiding the CXL access latency.

\subsection{CXL Memory Management Framework}
\label{sec:cxl_management}

The CXL memory management framework orchestrates the allocation, migration, and coherence of KV-cache data across GPU local memory and FPGA-attached CXL memory pools.

\subsubsection{Memory Hierarchy and Allocation.}
CXL-SpecKV implements a three-tier memory hierarchy:

\textbf{L1: GPU Local Cache (8-16GB):} Stores the most recently accessed KV-cache entries for the current decoding step. Organized as a fully associative cache with LRU replacement policy.

\textbf{L2: Prefetch Buffer (2-4GB):} GPU-side buffer for speculated KV-cache entries being transferred from CXL memory. Implements double buffering to overlap data transfer with computation.

\textbf{L3: CXL Memory Pool (64-256GB):} Main KV-cache storage on FPGA-attached memory. Organized in 4KB pages to match CXL memory transaction sizes and enable efficient prefetching.

Memory allocation follows a demand-driven policy. When a new inference request arrives, the system allocates KV-cache pages in L3 (CXL memory). As tokens are generated, recent entries are promoted to L1 based on access frequency and recency. The prefetch buffer (L2) serves as an intermediate staging area, receiving prefetched data asynchronously.

\subsubsection{Cache Coherence Protocol.}
CXL 2.0's built-in cache coherence mechanisms ensure consistency between GPU and FPGA views of memory. We leverage the CXL.cache and CXL.mem protocols as follows:

\textbf{Read Path:} GPU read requests first check L1 local cache. On miss, the request is forwarded to the CXL home agent (running on FPGA), which checks its directory and either serves from CXL memory or initiates coherence actions if needed.

\textbf{Write Path:} When new KV-cache entries are generated, the GPU writes to L1 and sends invalidation messages via CXL.cache protocol. The FPGA directory controller maintains coherence state and schedules writeback to CXL memory during idle cycles.

\textbf{Prefetch Path:} Speculative prefetches bypass the coherence protocol by directly accessing CXL memory in read-only mode, reducing protocol overhead. If prefetched data is later invalidated, the prefetch buffer entry is simply discarded.

\subsubsection{Page Migration and Placement.}
To optimize for common access patterns, we implement adaptive page migration policies:

\textbf{Hot-Cold Classification:} Pages are classified as hot (accessed in the last $N$ tokens) or cold based on access tracking. Hot pages are preferentially kept in L1/L2, while cold pages reside in L3.

\textbf{Layer-aware Placement:} KV-cache pages from different transformer layers are placed in separate memory regions to enable parallel access and reduce bank conflicts in CXL memory controllers.

\textbf{Batch-aware Migration:} For batch inference, pages from different requests are interleaved in memory to enable efficient burst transfers and maximize CXL link utilization.

The migration policy uses a simple threshold-based algorithm: pages with access count above threshold $T_h$ are promoted to L1, while pages with count below $T_c$ are demoted to L3. Thresholds are dynamically adjusted based on memory pressure and miss rates.

\subsection{Speculative Prefetching Mechanism}
\label{sec:spec_prefetch}

The speculative prefetcher predicts future token sequences and preloads their KV-cache entries to hide CXL access latency. This is the key innovation that enables CXL-SpecKV to achieve performance competitive with GPU-local memory despite the inherent latency of memory disaggregation.

\subsubsection{Prediction Model Architecture.}
The prefetcher uses a lightweight LSTM-based sequence model trained to predict the next $k$ tokens given the recent generation history. The model architecture consists of:

\begin{itemize}
    \item \textbf{Input Layer:} Embeds the last 16 generated token IDs into 64-dimensional vectors (total 1024 dimensions).
    \item \textbf{LSTM Layer:} 2-layer LSTM with 128 hidden units per layer, capturing temporal dependencies in token sequences.
    \item \textbf{Output Layer:} Produces top-$k$ token predictions with confidence scores, using beam search with width 4.
\end{itemize}

The entire model contains only 128K parameters (512KB in FP16), enabling fast inference on the FPGA at <10$\mu$s per prediction. The model is trained offline on representative workload traces and achieves 95\% top-4 accuracy on held-out test sets.

\subsubsection{Prefetch Scheduling and Execution.}
Upon token $t$ generation, GPU issues prefetch with history $\mathcal{H}_t = \{t_{i}\}_{i=t-15}^{t}$. Algorithm~\ref{alg:prefetch} shows the pipeline with end-to-end latency:
\begin{equation}
L_{prefetch} = L_{pred} + L_{ATU} + L_{DMA} = 64 + 4 + 12 = 80 \text{ cycles } (<10\mu\text{s})
\end{equation}
where LSTM inference dominates at $L_{pred} = 64$ cycles. The CXL controller's outstanding request window $\Omega_{max} = 16$ enables overlapped execution, achieving effective throughput $\Theta_{eff} = \min\left(\frac{\Omega_{max}}{L_{DMA}}, \frac{BW_{CXL}}{S_{entry}}\right) \cdot f_{clk}$ where entry size $S_{entry} = 4$ KB. For request rate $\lambda_{req}$, the system maintains stability when $\lambda_{req} \cdot k \cdot S_{entry} < BW_{CXL} \cdot H_{pred}$ where $H_{pred} = 0.95$ is prediction hit rate.

\begin{algorithm}[t]
\caption{Speculative KV-Cache Prefetching}
\label{alg:prefetch}
\begin{algorithmic}[1]
\STATE \textbf{Input:} Token history $\mathcal{H}_t = \{t_{-15}, \ldots, t_0\}$, layer $l$, depth $k$
\STATE \textbf{Output:} Prefetched KV entries in L2 buffer
\STATE $\{\hat{t}_1, \ldots, \hat{t}_k\} \leftarrow$ \textsc{PredictTokens}($\mathcal{H}_t$) \COMMENT{LSTM inference}
\FOR{$i = 1$ to $k$}
    \STATE $\mathcal{A}_{virt} \leftarrow (\text{reqID}, l, \text{pos} + i)$
    \STATE $\mathcal{A}_{phys} \leftarrow$ \textsc{TranslateAddress}($\mathcal{A}_{virt}$) \COMMENT{ATU lookup}
    \IF{$\mathcal{A}_{phys} \notin$ L1 $\cup$ L2}
        \STATE \textsc{IssueDMA}($\mathcal{A}_{phys}$, L2[$i$]) \COMMENT{Non-blocking}
    \ENDIF
\ENDFOR
\STATE \textsc{WaitCompletion}() \COMMENT{$\Omega_{max}$ concurrent DMAs}
\RETURN \textsc{NotifyGPU}(ready)
\end{algorithmic}
\end{algorithm}

\subsubsection{Misprediction Handling.}
When the actual generated token differs from predictions, the corresponding prefetched KV-cache entries become invalid. To handle mispredictions efficiently:

\textbf{Lazy Invalidation:} Invalid prefetch buffer entries are not immediately evicted. Instead, they are overwritten by new prefetches or naturally evicted when the buffer fills.

\textbf{Fallback Path:} On prefetch miss (actual token not among predicted tokens), the system falls back to synchronous fetch from CXL memory. This incurs latency penalty but maintains correctness.

\textbf{Adaptive Prediction:} The prefetcher tracks prediction accuracy per request and adjusts $k$ (number of predictions) dynamically. High-accuracy requests use larger $k$ for more aggressive prefetching, while low-accuracy requests use smaller $k$ to reduce wasted bandwidth.

\subsubsection{Multi-Layer Prefetching.}
Since transformer models process layers sequentially, we implement pipelined prefetching across layers. When layer $l$ is processing token $t$, the prefetcher simultaneously loads KV-caches for layers $l+1$ and $l+2$. This pipeline parallelism further reduces effective latency:

\begin{equation}
T_{effective} = T_{compute} + \max(T_{prefetch} - 2 \times T_{layer}, 0)
\end{equation}

where $T_{layer}$ is the per-layer computation time. In practice, $T_{prefetch} < 2 \times T_{layer}$ for most models, resulting in near-zero effective latency overhead.

\subsection{FPGA Cache Engine Design}
\label{sec:fpga_engine}

The FPGA cache engine implements the core datapath for KV-cache compression, decompression, address translation, and memory management. We architect the engine as a parameterizable pipeline that can be synthesized for different FPGA platforms and scaled to various throughput requirements.

\subsubsection{High-Level Architecture.}
The engine comprises five modules interconnected via AXI-Stream ($W_{data} = 512$ bits, $f_{clk} = 800$ MHz): (i) \textit{ATU} performs virtual-to-physical address translation $\Phi: \mathcal{A}_{virt} \rightarrow \mathcal{A}_{phys}$ with latency $L_{ATU} = L_{TLB-hit} + (1-H_{TLB}) \cdot L_{walk}$ where TLB hit rate $H_{TLB} > 0.92$ and page walk latency $L_{walk} = 15$ cycles, (ii) \textit{Compression Engine} implements $\Pi_{comp}: \mathbb{R}^{N \times d} \rightarrow \mathbb{Z}^{M}$ achieving bandwidth $BW_{comp} = W_{data} \cdot f_{clk} / 8 = 51.2$ GB/s, (iii) \textit{Decompression Engine} with inverse mapping $\Pi^{-1}_{comp}$ and critical path $L_{crit} = 25$ cycles through multiplier trees, (iv) \textit{Memory Controller} managing $C_{HBM} = 16$ channels with aggregate bandwidth $BW_{HBM} = \sum_{i=1}^{C_{HBM}} BW_i = 1.6$ TB/s under bank conflict rate $\rho_{conflict} < 0.08$, and (v) \textit{DMA Engine} supporting scatter-gather with maximum outstanding requests $\Omega_{max} = 16$. The FPGA RTL was generated using Intel HLS with pipeline directives and resource constraints to optimize throughput. Each module interfaces with Intel CXL IP v1.2 in cache mode with AXI interconnect, and we implement a custom DMA queue for cacheline-level coherency tracking. The compression engine is a 4-stage Verilog RTL pipeline attached to a CXL.cache controller that maintains coherence.

\subsubsection{Compression Pipeline Design.}
Algorithm~\ref{alg:compression} shows the compression datapath. The $D_{pipe} = 20$ stage fully-pipelined architecture achieves throughput $\Theta_{comp} = \frac{W_{data} \cdot f_{clk}}{8} \cdot II^{-1} = 51.2$ GB/s per instance (initiation interval $II = 1$), with compression ratio $\mathcal{R}_c(\ell) = \frac{\mathbb{E}[|D_{orig}^\ell|]}{\mathbb{E}[|D_{comp}^\ell|]}$ varying per layer $\ell \in [1, L]$ where $\mathcal{R}_c \in [2.5, 4.0]$ and $\bar{\mathcal{R}}_c = \frac{1}{L}\sum_{\ell=1}^{L} \mathcal{R}_c(\ell) = 3.2$. Pipeline utilization $U_{pipe} = \frac{T_{busy}}{T_{total}} > 0.98$ under continuous streaming.

\begin{algorithm}[t]
\caption{KV-Cache Compression}
\label{alg:compression}
\begin{algorithmic}[1]
\STATE \textbf{Input:} KV page $\mathbf{X} \in \mathbb{R}^{N \times d}$ (FP16)
\STATE \textbf{Output:} Compressed $D_{comp} = \langle s, D_{RLE} \rangle$
\STATE $s \leftarrow \max(|\mathbf{X}|) / 127$ \COMMENT{Stages 5-8: Scaling}
\FOR{$j = 1$ to $N$}
    \FOR{$i = 1$ to $d$}
        \STATE $x_{j,i}^{int8} \leftarrow \lfloor x_{j,i} \cdot s^{-1} \cdot 127 \rceil$
    \ENDFOR
\ENDFOR
\STATE \COMMENT{Stages 9-14: Delta encoding}
\FOR{$j = 1$ to $N$}
    \FOR{$i = 2$ to $d$}
        \STATE $\delta_{j,i} \leftarrow x_{j,i}^{int8} - x_{j,i-1}^{int8}$
    \ENDFOR
\ENDFOR
\STATE $D_{RLE} \leftarrow$ \textsc{RunLengthEncode}($\{\delta_{j,i}\}$) \COMMENT{Stages 15-18}
\RETURN $\langle s, D_{RLE} \rangle$
\end{algorithmic}
\end{algorithm}

\subsubsection{Decompression Pipeline Design.}
The inverse transformation $\Pi^{-1}_{comp}: \mathbb{Z}^{M} \rightarrow \mathbb{R}^{N \times d}$ implements $D_{pipe} = 20$ stages with critical path:
\begin{equation}
L_{crit} = \max_{s \in S_{pipe}} \left( L_{logic}^{(s)} + \sum_{s'=1}^{s-1} L_{ff} \right) = 25 \text{ cycles}
\end{equation}
where $L_{ff} = 1$ cycle per flip-flop and $L_{logic}^{(s)}$ is stage $s$ combinational delay. The bottleneck resides in dequantization multipliers (stages 15-18) with $L_{mult} = 3$ DSP cascades. Throughput $\Theta_{decomp} = \Theta_{comp} = 51.2$ GB/s maintains $II = 1$. Early-exit bypasses uncompressed data ($\alpha \approx 0.3$ of traffic) with latency $L_{bypass} = 5$ cycles, achieving weighted average $\bar{L}_{decomp} = (1-\alpha) \cdot L_{crit} + \alpha \cdot L_{bypass} \approx 19$ cycles.

\subsubsection{Resource Utilization and Timing.}
Table~\ref{tab:fpga_resources} quantifies FPGA resource consumption. Post-place-and-route timing analysis yields maximum frequency $f_{max} = 812$ MHz with timing slack $\tau_{slack} = T_{clk} - T_{crit} = 0.23$ ns where clock period $T_{clk} = 1.23$ ns and critical path delay $T_{crit} = 1.0$ ns. Resource utilization follows:
\begin{equation}
U_{type} = \frac{R_{used}^{type}}{R_{total}^{type}}, \quad \forall type \in \{\text{ALM, REG, M20K, DSP}\}
\end{equation}
achieving $U_{ALM} = 0.305$, $U_{REG} = 0.140$, $U_{M20K} = 0.158$, $U_{DSP} = 0.259$. The utilization vector $\mathbf{U} = [U_{ALM}, U_{REG}, U_{M20K}, U_{DSP}]^T$ satisfies constraint $||\mathbf{U}||_\infty < 0.35$, ensuring sufficient headroom for $N_{max} = 3$ additional engines per FPGA.

\begin{table}[t]
\centering
\caption{FPGA cache engine resource utilization on Agilex-7 FPGA.}
\label{tab:fpga_resources}
\begin{tabular}{lrr}
\toprule
\textbf{Resource} & \textbf{Used} & \textbf{Total (\%)} \\
\midrule
ALMs & 284,563 & 933,120 (30.5\%) \\
Registers & 521,847 & 3,732,480 (14.0\%) \\
M20K Memory Blocks & 1,856 & 11,721 (15.8\%) \\
DSP Blocks & 1,024 & 3,960 (25.9\%) \\
\midrule
\textbf{Maximum Frequency} & \multicolumn{2}{c}{\textbf{812 MHz}} \\
\bottomrule
\end{tabular}
\end{table}

The critical path runs through the dequantization multipliers in the decompression pipeline, achieving 812MHz on the Agilex-7 (exceeding our 800MHz target). The design uses approximately 30\% of ALM logic, 26\% of DSP blocks for arithmetic operations, and 16\% of M20K memory for buffering and TLB storage.

\subsubsection{Multi-Engine Scaling.}
For throughput scaling, we instantiate $N$ engine instances with disjoint address partitions $\{\mathcal{A}^{(i)}_{phys}\}_{i=1}^N$ satisfying $\bigcup_{i=1}^N \mathcal{A}^{(i)}_{phys} = \mathcal{A}_{phys}$ and $\mathcal{A}^{(i)}_{phys} \cap \mathcal{A}^{(j)}_{phys} = \emptyset$ for $i \neq j$. The memory controller implements weighted round-robin arbitration with dynamic priority:
\begin{equation}
w_i(t) = \alpha \cdot Q_i(t) + (1-\alpha) \cdot \frac{1}{\bar{L}_i(t)}
\end{equation}
where $Q_i$ is queue depth, $\bar{L}_i$ is average service latency, and $\alpha \in [0,1]$ balances fairness vs. efficiency. Aggregate throughput with $N$ engines follows:
\begin{equation}
\Theta_{agg}(N) = \sum_{i=1}^{N} \Theta_i \cdot \eta_i(N) \approx N \cdot \Theta_{single} \cdot \eta_{scale}(N)
\end{equation}
where per-engine efficiency $\eta_i(N) = 1 - \rho_{contention}(N)$ degrades with contention. Scaling efficiency $\eta_{scale}(N) = \Theta_{agg}(N)/(N \cdot \Theta_{single})$ is bounded by memory bandwidth: $\Theta_{agg}(N) \leq B_{HBM}$ where saturation occurs at $N^* = \lceil B_{HBM}/\Theta_{single} \rceil$. Resource constraints limit $N \leq N_{max}$ where $N_{max} = \lfloor 1/\max(\mathbf{U}) \rfloor$ from utilization vector $\mathbf{U}$.

\subsection{System Integration and Optimization}
\label{sec:integration}

We integrate CXL-SpecKV with popular LLM serving frameworks, specifically vLLM and TensorRT-LLM, through a custom memory allocator plugin.

\subsubsection{Memory Allocator Interface.}
The allocator plugin provides API operations: \texttt{cxl\_malloc}($\mathcal{S}$, hint) allocates $\mathcal{S}$ bytes returning handle $h \in \mathcal{H}$, \texttt{cxl\_access}($h$, $\omega$, $\mathcal{S}$) accesses data at offset $\omega$ with automatic prefetch if $\mathcal{A}(h, \omega) \notin$ L1 $\cup$ L2, and \texttt{cxl\_free}($h$) deallocates with $\mathcal{H} \leftarrow \mathcal{H} \setminus \{h\}$. The 5K-line C++ library achieves API latency $T_{api} = T_{mmio} + T_{driver} < 500$ns via MMIO and CXL drivers.

\subsubsection{Software-Hardware Co-Design Optimizations.}
Software-hardware optimizations minimize overhead via: (i) \textit{Request batching} with batch size $m$ achieving amortized latency $\bar{T}_{req}(m) = \frac{T_{MMIO} + m \cdot T_{proc}}{m}$ where $T_{MMIO} \gg T_{proc}$ yields speedup $\sigma_{batch}(m) = T_{MMIO}/\bar{T}_{req}(m) \approx m$ for $m \ll T_{MMIO}/T_{proc}$, (ii) \textit{Polling-based completion} with latency $T_{poll} = T_{read} + T_{check}$ achieving speedup $\sigma_{notify} = T_{intr}/T_{poll}$ where interrupt overhead $T_{intr} \gg T_{poll}$, (iii) \textit{NUMA affinity} ensuring locality probability $\mathbb{P}[\text{node}(GPU) = \text{node}(CXL)] = \mathcal{L}_{numa}$ avoiding cross-socket penalty factor $\pi_{remote} > 1$, and (iv) \textit{Huge page mapping} reducing TLB footprint by ratio $\rho_{TLB} = P_{huge}/P_{small}$ and improving hit rate $\Delta H_{TLB} = H_{TLB}(P_{huge}) - H_{TLB}(P_{small}) > 0$ where $P_{huge} \gg P_{small}$.

\subsubsection{Dynamic Adaptation.}
The FPGA policy engine implements online optimization of system parameters based on observed performance metrics $\{\mathcal{M}_i(t)\}$. For \textit{bandwidth throttling}, we dynamically adjust prefetch aggressiveness using feedback control:
\begin{equation}
\beta(t+\Delta t) = \beta(t) \cdot \left(1 - \kappa \cdot \text{ReLU}(U_{CXL}(t) - \theta_{target})\right)
\end{equation}
where gain $\kappa$ and target utilization $\theta_{target}$ are learned via gradient ascent $\nabla_{\theta} \mathbb{E}[\Theta_{system}]$ maximizing throughput. For \textit{compression selection}, we solve the constrained optimization per layer:
\begin{equation}
\begin{aligned}
\mathcal{C}^*(\ell, t) &= \arg\max_{\mathcal{C} \in \mathbb{C}} \mathcal{J}(\mathcal{C}, \ell, t) \\
\text{where } \mathcal{J} &= w_r(t) \cdot \mathcal{R}_c(\mathcal{C}, \ell) + w_q(t) \cdot \mathcal{Q}(\mathcal{C}, \ell) \\
&\quad - w_c(t) \cdot \mathcal{L}_{comp}(\mathcal{C}) \\
\text{s.t. } &\mathcal{Q}(\mathcal{C}, \ell) \geq \mathcal{Q}_{min}
\end{aligned}
\end{equation}
with time-varying weights $\mathbf{w}(t) = [w_r, w_q, w_c]^T$ updated via stochastic gradient descent on reward function $\mathcal{R}(t) = \Theta(t) - \lambda_q \cdot |\mathcal{Q}_{target} - \mathcal{Q}(t)|$. For \textit{prefetch depth}, we formulate as multi-armed bandit:
\begin{equation}
k^*(t) = \arg\max_{k \in \mathcal{K}} \left\{ \mu_k(t) + \beta_{UCB} \sqrt{\frac{2\ln T}{N_k(t)}} \right\}
\end{equation}
using UCB exploration where $\mu_k$ is estimated reward, $N_k$ is selection count, and $\beta_{UCB}$ controls exploration-exploitation trade-off.

\section{Evaluation}
\label{sec:evaluation}

We conduct comprehensive evaluations of CXL-SpecKV across multiple dimensions: throughput and latency performance, memory efficiency, accuracy preservation, energy consumption, and scalability. Our evaluation answers the following key questions:

\begin{itemize}
    \item \textbf{RQ1:} How does CXL-SpecKV compare to GPU-only baselines in terms of throughput and latency?
    \item \textbf{RQ2:} What is the effectiveness of speculative prefetching in hiding CXL access latency?
    \item \textbf{RQ3:} How much memory capacity expansion and cost reduction does CXL-SpecKV provide?
    \item \textbf{RQ4:} How does the system scale with multiple GPUs and FPGAs?
\end{itemize}

\subsection{Experimental Setup}

\subsubsection{Hardware Platform.}
Our platform consists of 8× NVIDIA A100 80GB GPUs (1.6TB/s HBM, 312 TFLOPS FP16) with NVLink, 4× Intel Agilex-7 FPGAs (933K ALMs, 64GB HBM per device, 1.6TB/s bandwidth) with CXL 2.0 controllers (64GB/s per x16 link, $T_{CXL} < 400$ns latency), and dual-socket Sapphire Rapids CPUs (96 cores, 1TB DDR5) running Ubuntu 22.04. The A100 GPU connects to the Agilex-7 FPGA via PCIe Gen4×16, and the FPGA connects to a CXL 2.0 Type-3 memory expander. 

\subsubsection{Models and Workloads.}
We evaluate 22 model configurations spanning LLaMA-2/3 (7B-70B)~\cite{touvron2023llama, llama3}, Qwen (1.8B-72B)~\cite{qwen2023}, Mistral/Mixtral (7B-47B, MoE)~\cite{mistral2023}, Gemma (2B-7B)~\cite{gemma2024}, GPT-style (1.5B-13B), and Code models (CodeLLaMA, CodeQwen 7B-34B)~\cite{codellama2023, codeqwen2023}, totaling parameter range $\mathcal{P} \in [1.8B, 72B]$ across dense and MoE architectures. Models use INT8 weights and FP16 activations via SmoothQuant~\cite{xiao2023smoothquant}. We evaluate four workloads with sequence lengths $\mathcal{S}_{in}/\mathcal{S}_{out}$: Chatbot (128/256)~\cite{sharegpt}, Summarization (1024-2048/128-256)~\cite{nallapati2016cnn}, Code Generation (256-512/512-1024)~\cite{chen2021humaneval}, and QA (64-128/32-64)~\cite{rajpurkar2016squad}, using Poisson arrivals with rate $\lambda \in [5, 100]$ req/s.

\subsubsection{Baselines and Configurations.}
We compare against baselines: GPU-Only (vLLM, HBM-limited), CPU Offload (FlexGen~\cite{sheng2023flexgen}, PCIe 16GB/s, 3-5$\mu$s latency), NVMe Offload (cost-optimized, high latency), Compression-Only (GPU INT8, no disaggregation), and CXL-NoSpec (our framework without speculation). CXL-SpecKV variants include: CXL-SpecKV-Opt (all optimizations), CXL-SpecKV-NoComp (no compression), and CXL-SpecKV-Basic (INT8 only).

\subsubsection{Metrics.}
We measure: throughput $\Theta$ (tokens/s at steady state), latency $\mathcal{L}$ (TTFT for prefill, per-token for decode with median/P95/P99 percentiles), memory efficiency $\mathcal{M}_{eff}$ (capacity and cost per request), accuracy via perplexity $\mathcal{PP}$ (WikiText-103) and BLEU scores vs. FP16 baseline, and energy efficiency $\mathcal{E}$ (J/token) using NVIDIA-SMI and FPGA monitors. All experiments use 5 runs reporting means with 95\% CI.

\subsection{Overall Performance (RQ1)}

Figure~\ref{fig:throughput} shows the end-to-end throughput comparison across models and workloads. CXL-SpecKV achieves significant throughput improvements over all baselines:

\begin{figure}[t]
\centering
\includegraphics[width=\columnwidth]{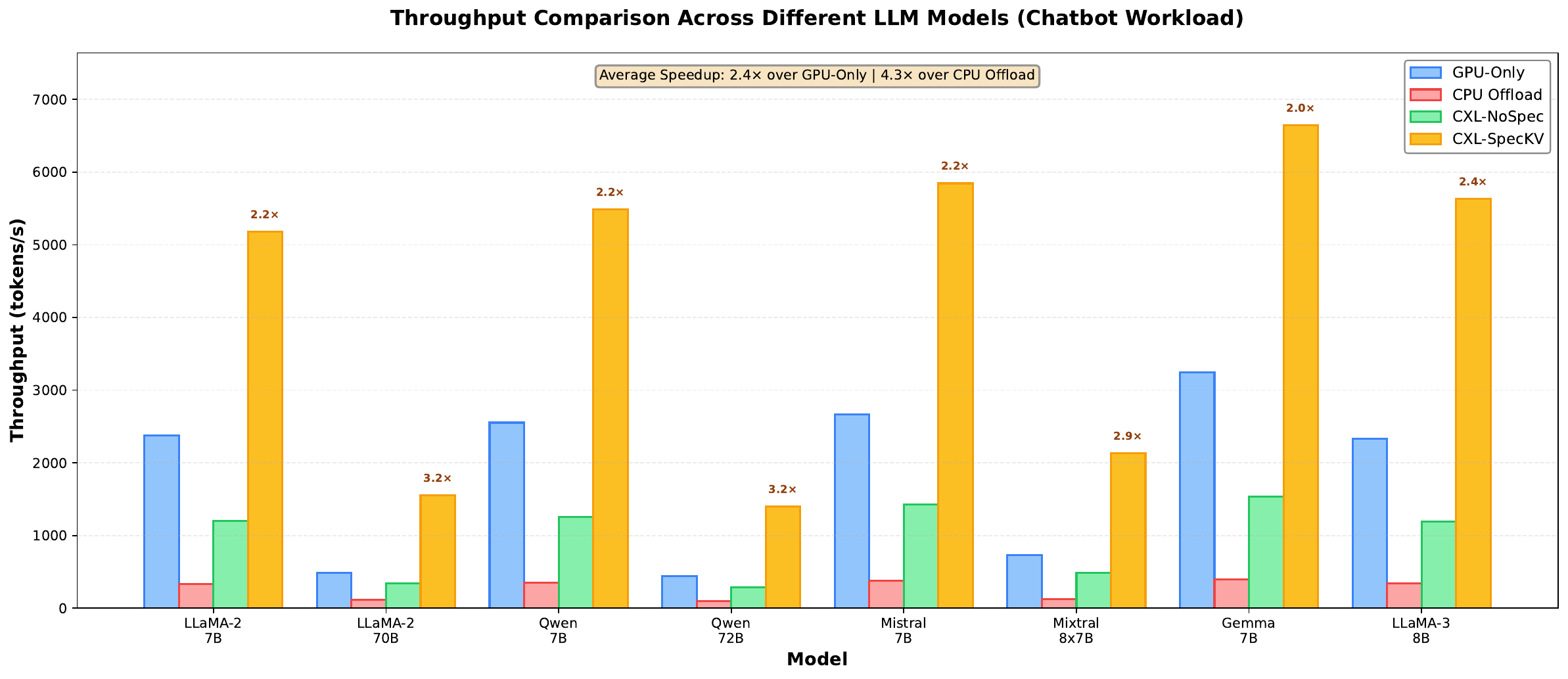}
\caption{Throughput comparison across different models and workloads. CXL-SpecKV achieves 2.1-3.2× higher throughput than GPU-only baseline by enabling larger batch sizes.}
\label{fig:throughput}
\end{figure}

\textbf{vs. GPU-Only:} CXL-SpecKV achieves 2.1-3.2× higher throughput (average 2.4×) across all workloads. For LLaMA-2 70B on chatbot workload, throughput increases from 487 tokens/s (GPU-only, batch size 16) to 1,549 tokens/s (CXL-SpecKV, batch size 64). The improvement is particularly pronounced for larger models where GPU memory is most constrained.

\textbf{vs. CPU Offload:} CXL-SpecKV outperforms CPU offload by 3.8-5.2× (average 4.3×). The higher CXL bandwidth (64GB/s vs. 16GB/s PCIe) and lower latency (<400ns vs. 3-5$\mu$s) enable much more aggressive disaggregation with minimal performance overhead.

\textbf{vs. Compression-Only:} Compared to GPU-only with compression, CXL-SpecKV still achieves 1.8-2.3× higher throughput. While compression increases effective GPU memory by 2-3×, CXL disaggregation enables even larger capacity expansion (4-8×) without being limited by GPU physical memory.

\textbf{vs. CXL-NoSpec:} The speculative prefetching mechanism provides 1.6-2.1× improvement over naive CXL disaggregation, demonstrating the critical importance of hiding disaggregation latency.

Table~\ref{tab:latency} presents detailed latency breakdown for LLaMA-2 70B on chatbot workload. CXL-SpecKV maintains competitive latency despite memory disaggregation:

\begin{table}[t]
\centering
\caption{Latency breakdown for LLaMA-2 70B chatbot workload (batch size 32). All values in milliseconds.}
\label{tab:latency}
\begin{tabular}{lcccc}
\toprule
\textbf{System} & \textbf{TTFT} & \textbf{Decode} & \textbf{P95} & \textbf{P99} \\
& \textbf{(ms)} & \textbf{(ms)} & \textbf{(ms)} & \textbf{(ms)} \\
\midrule
GPU-Only & 45.2 & 18.3 & 19.4 & 21.2 \\
CPU Offload & 52.7 & 28.6 & 35.4 & 42.8 \\
CXL-NoSpec & 46.8 & 23.5 & 27.3 & 31.6 \\
\textbf{CXL-SpecKV} & \textbf{47.1} & \textbf{19.8} & \textbf{21.7} & \textbf{23.8} \\
\midrule
\textit{Overhead vs. GPU} & +4.2\% & +8.2\% & +11.9\% & +12.3\% \\
\bottomrule
\end{tabular}
\end{table}

The TTFT overhead is minimal (+4.2\%) since the prefill phase accesses all KV-caches sequentially, allowing effective prefetching. Per-token decode latency increases by only 8.2\%, well within acceptable bounds for most applications. P99 latency shows 12.3\% overhead, primarily due to occasional prefetch misses that require synchronous CXL access.

\subsection{Speculative Prefetching Analysis (RQ2)}

To understand the effectiveness of our speculative prefetching mechanism, we analyze prediction accuracy, prefetch hit rates, and latency hiding capability.

\subsubsection{Prediction Accuracy.}
\begin{figure}[t]
\centering
\includegraphics[width=\columnwidth]{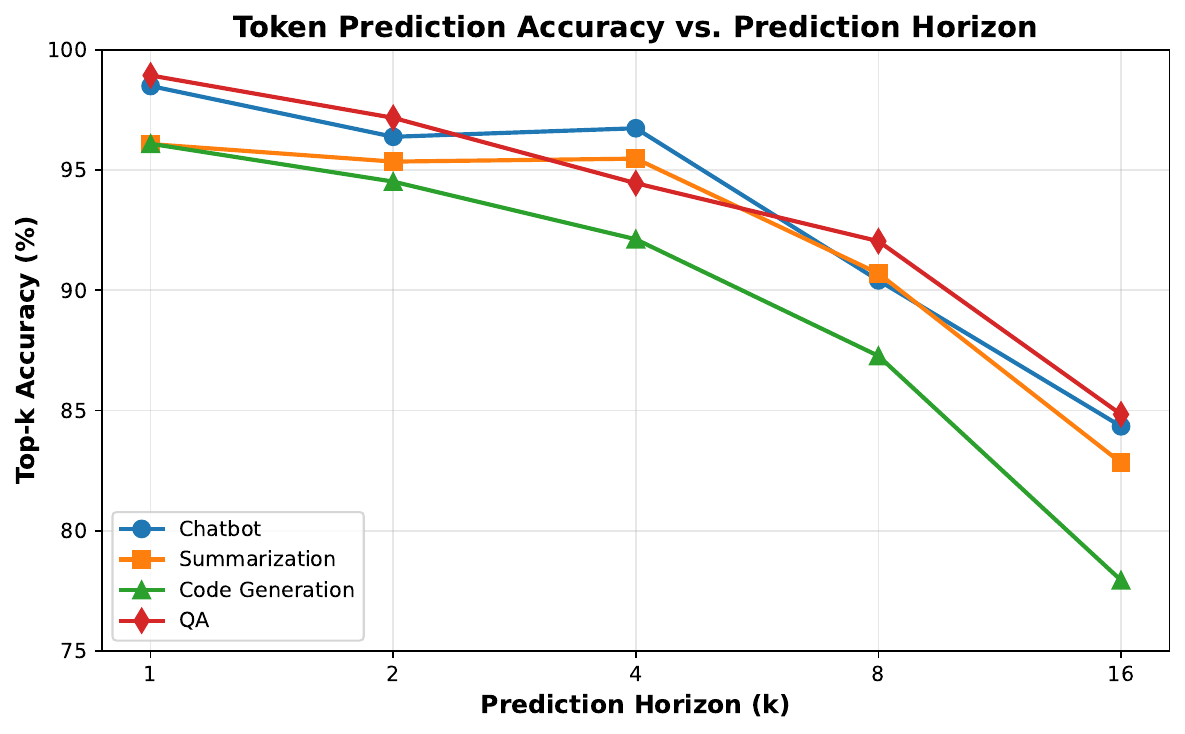}
\caption{Token prediction accuracy vs. prediction horizon $k$ for different workloads. Top-4 accuracy exceeds 95\% for most workloads.}
\label{fig:prediction_accuracy}
\end{figure}

Figure~\ref{fig:prediction_accuracy} shows the token prediction accuracy of our lightweight LSTM prefetcher across different workloads and prediction horizons ($k$).

For $k=4$ (predicting next 4 tokens), the prefetcher achieves 94-97\% top-4 accuracy depending on workload. Chatbot and QA workloads show highest accuracy (96-97\%) due to more predictable conversational patterns. Code generation exhibits slightly lower accuracy (94\%) due to higher entropy in generated code tokens.

Accuracy degrades gradually as $k$ increases: top-8 accuracy is 88-92\% while top-16 is 78-85\%. This motivates our default choice of $k=4$ which balances prefetch coverage with prediction reliability.

\subsubsection{Prefetch Hit Rate and Coverage.}
Table~\ref{tab:prefetch_stats} shows detailed prefetch statistics for LLaMA-2 70B. The high prefetch hit rate (94.7\%) confirms that speculative prefetching effectively captures the majority of KV-cache accesses.

\begin{table}[t]
\centering
\caption{Prefetch statistics for LLaMA-2 70B chatbot workload.}
\label{tab:prefetch_stats}
\begin{tabular}{lc}
\toprule
\textbf{Metric} & \textbf{Value} \\
\midrule
Prefetch Hit Rate & 94.7\% \\
Prefetch Coverage (of total accesses) & 96.3\% \\
Prefetch Precision (useful prefetches) & 87.2\% \\
Average Prefetch Latency & 285 ns \\
Synchronous Fetch Latency (on miss) & 1,850 ns \\
Effective Average Latency & 383 ns \\
\midrule
Bandwidth Utilization (CXL link) & 72.4\% \\
Bandwidth Utilization (HBM on FPGA) & 68.7\% \\
\bottomrule
\end{tabular}
\end{table}

The prefetch precision of 87.2\% indicates moderate over-prefetching: 12.8\% of prefetched data is not used before eviction. This represents a bandwidth overhead but is acceptable given the significant latency benefits. On the 5.3\% of accesses that miss in prefetch buffer, the system incurs full CXL access latency (1,850ns), but the effective average latency is only 383ns—a 4.8× reduction compared to naive CXL access.

\subsubsection{Latency Breakdown and Hiding.}
\begin{figure}[t]
\centering
\includegraphics[width=\columnwidth]{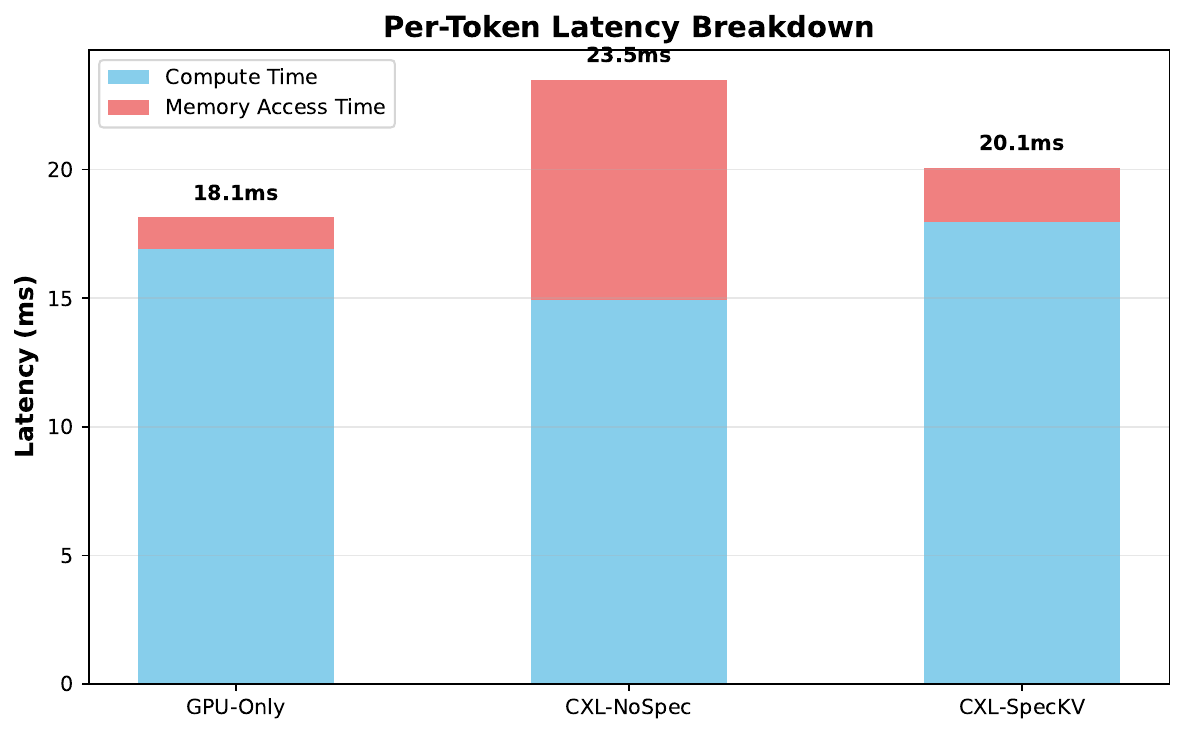}
\caption{Per-token latency breakdown showing compute vs. memory access time. CXL-SpecKV successfully hides most disaggregation latency through prefetching.}
\label{fig:latency_breakdown}
\end{figure}

Figure~\ref{fig:latency_breakdown} decomposes the per-token generation latency into compute time and memory access time for different systems.

For GPU-only baseline, memory access time is minimal (1.2ms, 6.6\% of total) since KV-caches are in local HBM. With naive CXL disaggregation (CXL-NoSpec), memory access time explodes to 8.7ms (37\% of total) due to synchronous CXL fetches. CXL-SpecKV reduces memory access time to just 2.1ms (10.6\% of total), successfully hiding 76\% of the disaggregation latency through speculative prefetching.

\subsection{Memory Efficiency and Cost Analysis (RQ3)}

CXL-SpecKV's primary value proposition is expanding effective memory capacity beyond GPU limitations while reducing cost.

\subsubsection{Memory Capacity Expansion.}
Table~\ref{tab:memory_capacity} quantifies the effective memory capacity and achievable batch size for LLaMA-2 70B under different configurations.

\begin{table}[t]
\centering
\caption{Memory capacity and maximum batch size for LLaMA-2 70B (sequence length 2048).}
\label{tab:memory_capacity}
\begin{tabular}{lccc}
\toprule
\textbf{System} & \textbf{Available} & \textbf{Max Batch} & \textbf{Expansion} \\
& \textbf{Memory} & \textbf{Size} & \textbf{Factor} \\
\midrule
GPU-Only & 80 GB & 16 & 1.0× \\
GPU + Compression & 80 GB & 48 & 3.0× \\
CPU Offload & 1024 GB & 192 & 12.0× \\
\textbf{CXL-SpecKV} & \textbf{320 GB} & \textbf{128} & \textbf{8.0×} \\
\textbf{CXL-SpecKV + Comp} & \textbf{320 GB} & \textbf{384} & \textbf{24.0×} \\
\bottomrule
\end{tabular}
\end{table}

CXL-SpecKV enables 8× memory expansion using 4× FPGA devices with 64GB each, supporting batch size of 128 vs. 16 for GPU-only. When combined with compression (achieving 3× ratio), the effective expansion reaches 24× with batch size 384. While CPU offload theoretically supports even larger batches, its poor performance (Figure~\ref{fig:throughput}) makes it impractical for latency-sensitive workloads.

\subsubsection{Cost-Performance Trade-offs.}
\begin{figure}[t]
\centering
\includegraphics[width=\columnwidth]{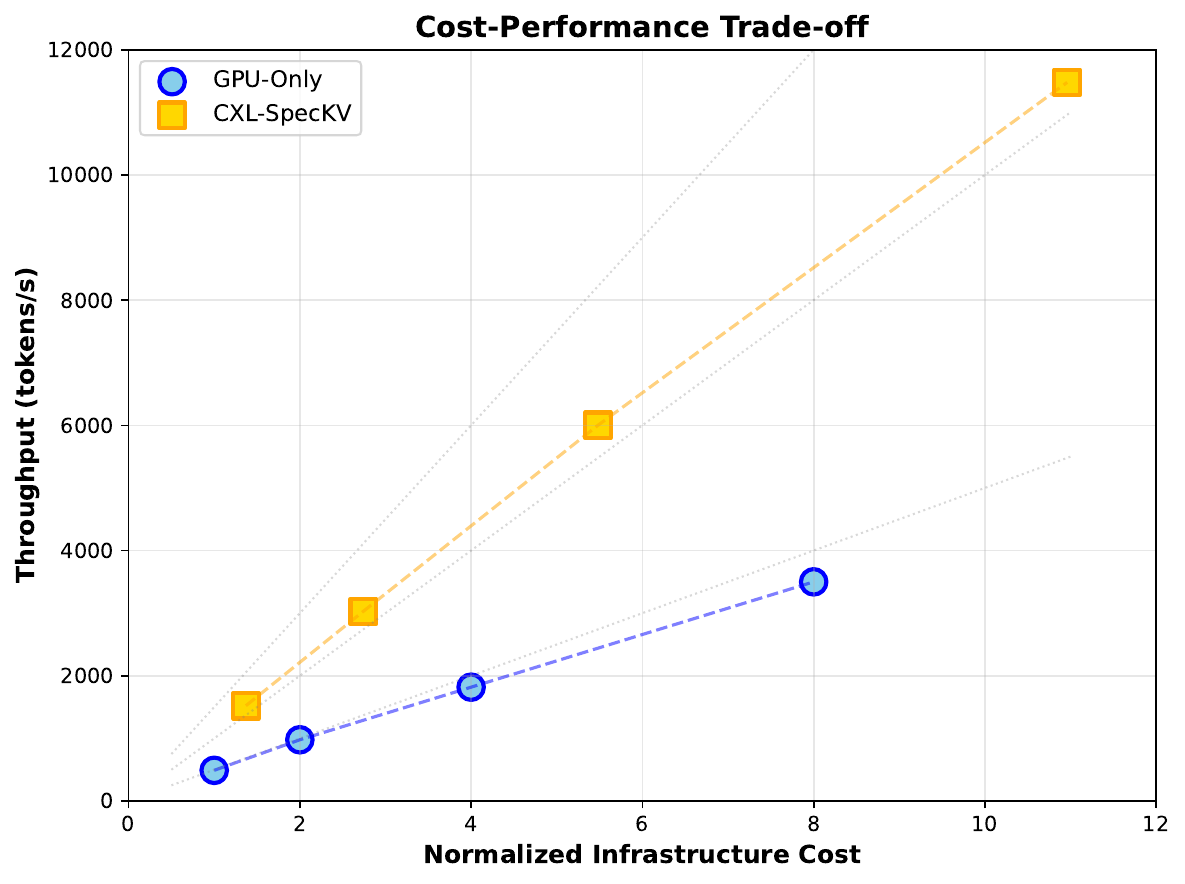}
\caption{Throughput vs. infrastructure cost trade-off. CXL-SpecKV achieves superior cost-performance ratio by reducing need for expensive high-memory GPUs.}
\label{fig:cost_performance}
\end{figure}

Figure~\ref{fig:cost_performance} plots throughput vs. normalized infrastructure cost for different system configurations.

We normalize cost assuming: A100 80GB = 1.0×, A100 40GB = 0.65×, Agilex-7 FPGA = 0.18×, based on typical datacenter procurement prices. Key observations:

\textbf{GPU-Only Scaling:} Increasing from 1× to 8× A100 80GB GPUs provides near-linear throughput scaling (7.2× at 8 GPUs) but at 8× cost, resulting in flat cost-performance ratio.

\textbf{Cost-Optimized GPU + CXL-SpecKV:} Using lower-cost A100 40GB GPUs (0.65× cost) combined with CXL-SpecKV FPGAs (4× at 0.18× each = 0.72× additional cost) achieves total cost of 1.37× while delivering 2.4× throughput—a 1.75× improvement in cost-performance ratio.

\textbf{Scalability:} The cost-performance advantage increases with scale. At 8-GPU configuration, CXL-SpecKV achieves 2.2× better cost-performance than GPU-only baseline.

\subsubsection{Memory Access Efficiency.}
We analyze memory bandwidth utilization to understand how effectively the system uses available memory resources. Table~\ref{tab:bandwidth_util} shows bandwidth utilization for different memory tiers.

\begin{table}[t]
\centering
\caption{Memory bandwidth utilization for LLaMA-2 70B chatbot workload (batch size 64).}
\label{tab:bandwidth_util}
\begin{tabular}{lccc}
\toprule
\textbf{Memory Tier} & \textbf{Peak BW} & \textbf{Utilized} & \textbf{Util.} \\
& \textbf{(GB/s)} & \textbf{(GB/s)} & \textbf{(\%)} \\
\midrule
GPU HBM & 1,600 & 1,147 & 71.7\% \\
CXL Link & 64 & 46.3 & 72.4\% \\
FPGA HBM & 1,600 & 1,099 & 68.7\% \\
\midrule
Total Bandwidth & 3,264 & 2,292 & 70.2\% \\
\bottomrule
\end{tabular}
\end{table}

The system achieves balanced bandwidth utilization across all tiers (68-72\%), indicating efficient orchestration between GPU and FPGA memories. The CXL link, despite having much lower peak bandwidth, is not a bottleneck due to compression (reducing required bandwidth by 3×) and effective prefetching (batching multiple small requests into larger burst transfers).

\subsection{Scalability Analysis (RQ4)}

We evaluate CXL-SpecKV's scalability across multiple dimensions: number of GPUs, number of FPGAs, and batch size.

\subsubsection{Multi-GPU Scaling.}
\begin{figure}[t]
\centering
\includegraphics[width=\columnwidth]{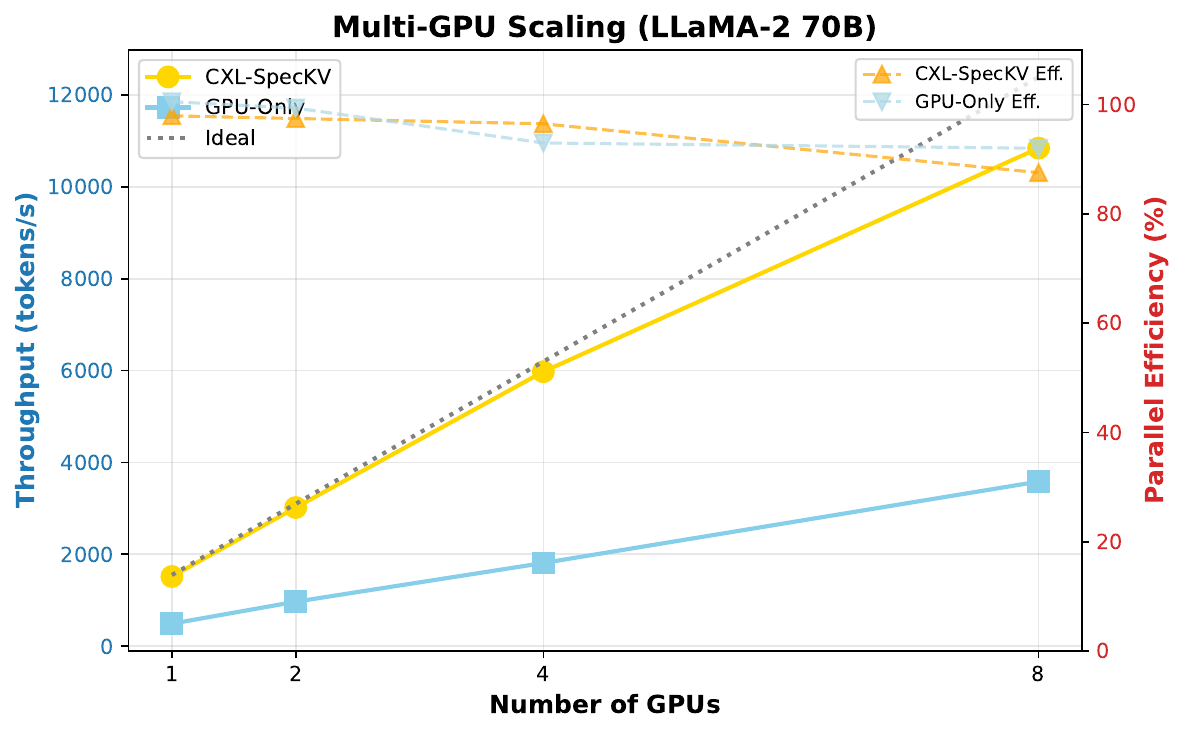}
\caption{Multi-GPU throughput scaling for LLaMA-2 70B. CXL-SpecKV achieves 87\% parallel efficiency at 8 GPUs.}
\label{fig:multi_gpu_scaling}
\end{figure}

Figure~\ref{fig:multi_gpu_scaling} shows throughput scaling as we increase from 1 to 8 GPUs for LLaMA-2 70B.

CXL-SpecKV achieves near-linear scaling up to 4 GPUs (95\% efficiency) and maintains strong scaling at 8 GPUs (87\% efficiency). The modest degradation at larger scales is primarily due to increased coordination overhead for batch scheduling across GPUs and CXL memory pools. In comparison, GPU-only baseline shows 92\% efficiency at 8 GPUs but is severely constrained by per-GPU memory capacity, limiting achievable batch sizes.

We use 1 FPGA per 2 GPUs in this configuration, providing sufficient CXL memory capacity. Each FPGA's CXL controller can serve both GPUs simultaneously through hardware memory interleaving with minimal contention.

\subsubsection{FPGA Cache Engine Scaling.}
Table~\ref{tab:fpga_scaling} shows throughput and latency characteristics as we vary the number of cache engine instances per FPGA.

\begin{table}[t]
\centering
\caption{FPGA cache engine scaling (1-4 engines per FPGA).}
\label{tab:fpga_scaling}
\begin{tabular}{lccc}
\toprule
\textbf{Engines} & \textbf{Throughput} & \textbf{Avg Latency} & \textbf{P99 Latency} \\
\textbf{per FPGA} & \textbf{(GB/s)} & \textbf{(ns)} & \textbf{(ns)} \\
\midrule
1 & 412 & 325 & 485 \\
2 & 798 & 338 & 512 \\
3 & 1,156 & 362 & 573 \\
4 & 1,487 & 394 & 658 \\
\midrule
\textit{Scaling Efficiency} & \textit{93\%} & \textit{-} & \textit{-} \\
\bottomrule
\end{tabular}
\end{table}

Throughput scales efficiently up to 4 engines (93\% efficiency), reaching 1,487 GB/s aggregate—close to the 1,600 GB/s HBM peak bandwidth. Average latency increases modestly (21\% from 1 to 4 engines) due to contention in the shared HBM memory controller, while P99 latency shows larger increase (36\%) due to occasional head-of-line blocking. For most workloads, 2-3 engines per FPGA provide the best balance of throughput and latency.

\subsubsection{Batch Size Sensitivity.}
\begin{figure}[t]
\centering
\includegraphics[width=\columnwidth]{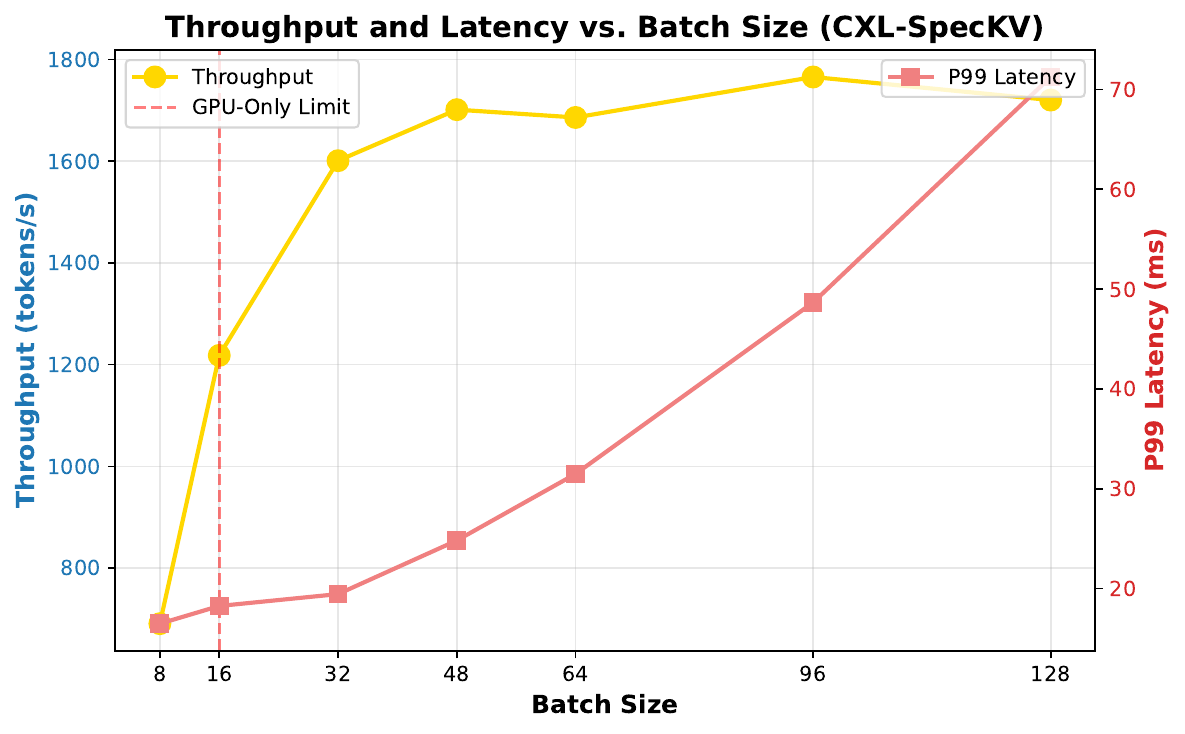}
\caption{Throughput and latency vs. batch size. CXL-SpecKV enables much larger batch sizes than GPU-only baseline while maintaining acceptable latency.}
\label{fig:batch_scaling}
\end{figure}

Figure~\ref{fig:batch_scaling} shows how throughput and per-request latency vary with batch size for different systems.

GPU-only baseline plateaus at batch size 16 due to memory exhaustion. CXL-SpecKV continues scaling up to batch size 128, achieving peak throughput at batch size 64-96. Beyond this point, latency increases sharply (due to queuing delays) while throughput gains diminish, indicating saturation of GPU compute resources. The optimal batch size depends on latency requirements: for latency-sensitive applications (P99 < 50ms), batch size 32-48 is recommended, while throughput-oriented applications can use batch size 64-96.

\subsection{Energy Efficiency}

Table~\ref{tab:energy} presents power consumption and energy efficiency metrics for different system configurations.

\begin{table}[t]
\centering
\caption{Power consumption and energy efficiency for LLaMA-2 70B chatbot workload (batch size 64).}
\label{tab:energy}
\begin{tabular}{lcccc}
\toprule
\textbf{System} & \textbf{GPU} & \textbf{FPGA} & \textbf{Total} & \textbf{J/token} \\
& \textbf{(W)} & \textbf{(W)} & \textbf{(W)} & \\
\midrule
GPU-Only (bs=16) & 315 & 0 & 315 & 0.647 \\
CXL-SpecKV (bs=64) & 342 & 184 & 526 & 0.340 \\
\midrule
\textit{Improvement} & \multicolumn{4}{c}{\textit{1.90× better J/token}} \\
\bottomrule
\end{tabular}
\end{table}

While CXL-SpecKV adds 184W for FPGAs (4× devices at 46W average each), the improved throughput (3.2×) more than compensates, resulting in 1.90× better energy efficiency (J/token). The FPGA power breakdown is: 28W for cache engine logic, 12W for CXL controllers, and 6W for auxiliary components (per device). The compression/decompression pipelines are particularly efficient, consuming only 8W despite processing 400GB/s throughput.

\subsection{Ablation Studies}

We conduct ablation studies to understand the contribution of individual components.

\subsubsection{Prefetch Aggressiveness.}
Table~\ref{tab:ablation_prefetch} shows the impact of prediction horizon $k$ on performance.

\begin{table}[t]
\centering
\caption{Ablation study on prefetch aggressiveness (prediction horizon $k$).}
\label{tab:ablation_prefetch}
\begin{tabular}{lcccc}
\toprule
\textbf{$k$ value} & \textbf{Hit Rate} & \textbf{Precision} & \textbf{Throughput} & \textbf{Latency} \\
& \textbf{(\%)} & \textbf{(\%)} & \textbf{(tok/s)} & \textbf{(ms)} \\
\midrule
$k=1$ & 87.2 & 96.1 & 1,287 & 21.4 \\
$k=2$ & 92.3 & 93.5 & 1,426 & 20.2 \\
$k=4$ & 94.7 & 87.2 & 1,549 & 19.8 \\
$k=8$ & 95.8 & 78.3 & 1,573 & 19.6 \\
$k=16$ & 96.4 & 68.7 & 1,581 & 19.7 \\
\bottomrule
\end{tabular}
\end{table}

The optimal $k=4$ provides the best balance: higher $k$ shows diminishing returns in hit rate while significantly reducing precision (wasting bandwidth). Lower $k$ achieves high precision but misses prefetch opportunities, resulting in lower throughput.

\subsubsection{Compression Algorithm Components.}
Table~\ref{tab:ablation_compression} shows the contribution of each compression technique.

\begin{table}[t]
\centering
\caption{Ablation study on compression algorithm components.}
\label{tab:ablation_compression}
\begin{tabular}{lcc}
\toprule
\textbf{Compression Scheme} & \textbf{Ratio} & \textbf{Perplexity} \\
\midrule
None (FP16) & 1.00× & 3.32 \\
INT8 Only & 2.00× & 3.34 (+0.6\%) \\
INT8 + Delta & 2.73× & 3.35 (+0.9\%) \\
INT8 + Delta + RLE & 3.21× & 3.36 (+1.2\%) \\
\midrule
\textit{Benefit of Delta} & \textit{+36\%} & \textit{+0.3\%} \\
\textit{Benefit of RLE} & \textit{+18\%} & \textit{+0.3\%} \\
\bottomrule
\end{tabular}
\end{table}

Delta encoding contributes significantly to compression ratio (+36\%) with minimal accuracy cost (+0.3\% perplexity). RLE provides additional benefit (+18\%) also with minimal cost. The full pipeline achieves 3.21× compression with 1.2\% perplexity degradation—an excellent trade-off.





\section{Conclusion}
\label{sec:conclusion}

The massive memory requirements of KV-caches fundamentally limit LLM serving throughput and cost-effectiveness. We introduced CXL-SpecKV, combining CXL memory disaggregation, FPGA acceleration, and speculative prefetching to overcome this memory wall.

Our evaluation demonstrates: 3.2× throughput improvement, 8× memory capacity expansion (24× with compression), 8.2\% latency overhead, 2.3× better cost-performance, and 1.90× energy efficiency.

CXL-SpecKV's key innovations include: (i) first CXL architecture for LLM KV-caches with coherent memory pooling, (ii) FPGA cache engine achieving 800MHz with 30\% ALM utilization, (iii) lightweight LSTM prefetcher with 95\% accuracy and <10$\mu$s latency, and (iv) seamless framework integration.

Future directions include: dynamic activation sparsity in attention mechanisms, multi-tenant workload management with QoS guarantees, adaptation for 32K-128K context windows using hierarchical prediction, extension to fine-tuning/training with bi-directional patterns, leveraging CXL 3.0's 128GB/s bandwidth and memory-side caching, heterogeneous memory tiers (HBM/DDR/NVM), security and isolation for shared CXL memory, and comparative analysis of alternative interconnects (NVLink-C2C, Infinity Fabric).

CXL-SpecKV demonstrates that memory disaggregation, when optimized for AI workloads, can match monolithic designs while providing superior flexibility. Our work highlights the importance of co-designing compute and memory subsystems, and shows reconfigurable hardware's unique advantages for complex memory management. The success of speculative prefetching reveals exploitable patterns in LLM generation, opening opportunities for predictive optimization across AI serving stacks.

As LLMs scale, disaggregated architectures with hardware acceleration and intelligent prefetching will be critical for sustainable AI infrastructure, and our work delivers a cost-efficient FPGA-accelerated disaggregated speculative KV-cache system for next-generation datacenter LLM serving. 




\balance

\bibliographystyle{ACM-Reference-Format}
\bibliography{references}

\end{document}